%% file: main.tex
\documentclass{article}

\usepackage{arxiv}
\RequirePackage{natbib}

\usepackage[utf8]{inputenc} 
\usepackage[T1]{fontenc}    
\usepackage{hyperref}       
\usepackage{url}            
\usepackage{booktabs}       
\usepackage{amsfonts}       
\usepackage{nicefrac}       
\usepackage{microtype}      
\usepackage[bottom]{footmisc} 
\usepackage{appendix}      

\usepackage{tikz}
\usetikzlibrary{decorations.pathreplacing,positioning, arrows.meta}

\title{MPLP: Learning a Message Passing Learning Protocol}

\author{%
  Ettore Randazzo \\
  Google Research \\
  \texttt{etr@google.com} \\
  \And
  Eyvind Niklasson \\
  Google Research \\
  \texttt{eyvind@google.com} \\
  \And
  Alexander Mordvintsev \\
  Google Research \\
  \texttt{moralex@google.com} \\
}

\begin{document}

\maketitle
\input{abstract_section}

\input{intro}
\input{relwork}
\input{model}
\section{Experiments}
\input{sinusoid}
\input{mnist}
\input{outro}

\section*{Acknowledgements}
We thank Iulia Comșa, Hugo Larochelle, Blaise Aguera y Arcas, and Max Vladymyrov for their insightful feedback.

\small
\bibliographystyle{plainnat}
\bibliography{main}

\input{appendix}

\end{document}

%% file: abstract_section.tex
\begin{abstract}

We present a novel method for learning the weights of an artificial neural network - a Message Passing Learning Protocol (MPLP). In MPLP, we abstract every operations occurring in ANNs as independent agents. Each agent is responsible for ingesting incoming multidimensional messages from other agents, updating its internal state, and generating multidimensional messages to be passed on to neighbouring agents. We demonstrate the viability of MPLP as opposed to traditional gradient-based approaches on simple feed-forward neural networks, and present a framework capable of generalizing to non-traditional neural network architectures. MPLP is meta learned using end-to-end gradient-based meta-optimisation. We further discuss the observed properties of MPLP and hypothesize its applicability on various fields of deep learning.

\end{abstract}

%% file: intro.tex
\section{Introduction}

For the most part, learning algorithms have been hand-crafted. The deep learning community has largely converged to use almost exclusively gradient-based approaches for learning a model's parameters. Such gradient-based approaches generally impose limitations in terms of the loss landscape, choice of network architecture and training dynamics. A non-exhaustive list of examples is: their inherent tendency to overfit to training sets, their catastrophic forgetting behaviour, their requirement of a smooth loss landscape, and experiencing vanishing or exploding gradients in recurrent or large architectures. Moreover, while the mechanics of artificial neurons are inspired by their biological counterparts, they have been greatly simplified to be scalable and differentiable, rendering them far less powerful than their biological counterpart. Perhaps the simplicity of its building blocks is the reason of why most of deep learning research occurs on layered deep networks that require increasing amounts of computation and memory, limiting the explorations on fundamentally different architectures. Nevertheless, backpropagation is still the best tool in our toolkit for optimising models with an extensive set of parameters.

In this work, we leverage gradient-based learning to find a new learning protocol for tuning an arbitrary computational graph to adapt to a task from a given family of tasks. We show how this learning protocol and its associated meta-learner can be used to train traditional neural networks. This is accomplished by rethinking neural networks as self-organising systems: a graph composed of nodes representing operations such as synapses (individual weights and biases), activations and losses, that have to communicate in order to solve a given task.

We therefore propose to learn a Message Passing Learning Protocol (MPLP): given a directed graph composed of (sparsely) connected nodes, we let these nodes communicate with each other by passing k-dimensional vectors along directed edges. The meta-training phase consists of learning a MPLP that, given an initial configuration/initialization, and a training set, is able to adapt to a given task. The kind of graphs we explore in this work are all end-to-end differentiable - we therefore meta-optimize MPLP through gradient-based approaches.

We show how MPLP can be applied to feed-forward neural networks as a meta-learned replacement to gradient-based approaches. In fact, we can consider gradient-based learning algorithms as a specific instance of MPLP. Gradient based approaches do the following: (1) in the forward pass, store the input $x$ and compute $y$ with your function. (2) in the backward pass, get a gradient of a loss upstream, compose it with your gradient function, and update your own weight with the gradient and some autoregressive behavior, (3) pass a gradient of the loss, modified accordingly to your function. Hence, we can specialize MPLP as follows: every time an operation occurs, we consider it a node. Examples of such nodes are single weights $w_{ij}$ (or biases $b_i$) multiplied by an input $x_i$, or a value modified by an activation function. The forward pass remains unchanged. In the backward pass, instead of receiving the simple gradient of a loss scalar, each node receives a multidimensional message vector, updates its weight through a parameterized function $f$, and backpropagates a modified message through a parameterized function $g$. These functions are trainable neural networks. Given that MLPs have been shown to work as universal function approximators, we propose this property would allow our learning protocol to implement traditional gradient descent arbitrarily well, but would no be limited to or encouraged to do exactly so. While for real-world applications it would be desirable to backpropagate the gradient of the loss alongside a learned message, we decided to \textit{never pass any gradient} to showcase properties of a pure MPLP. As a proof of concept, we show how a gradient-free MPLP can be used to train feedforward neural networks for few-shot sinusoidal fitting and MNIST classification. In the sinusoidal case, we also show how we can enhance Model-Agnostic Meta-Learning (MAML) (\citet{finn2017modelagnostic}) approaches by jointly learning both priors and learning rules.

While this paper limits the explorations on this framework to traditional feedforward neural networks, this framework can be used on any graph of agents, as long as their communication protocol remains end-to-end differentiable. We briefly discuss some possible applications of MPLP on non-traditional neural networks in Section~\ref{sec:discussion}. The code for this framework and reproduction of the following experiments can be found on \href{https://github.com/google-research/self-organising-systems/tree/master/mplp}{https://github.com/google-research/self-organising-systems/tree/master/mplp}.

%% file: relwork.tex
\section{Related Work}

\textbf{Meta-learning.} Our work finds extensive common ground in the field of meta-learning, which has recently re-exploded in popularity. Given the velocity and size of the field, we must limit this section to work we believe is very strongly related or has served as a direct inspiration to this work. For a more extensive overview of meta-learning, we recommend \citet{clune2019aigas} and \citet{hospedales2020metalearning}.

(\citet{bengio91, Bengio97onthe}) introduced the idea of discovering local learning rules instead of gradient-based optimization. We consider our work as a generalization of their approach.
\citet{schmidhuber1993} and follow-up work (\citet{Hochreiter01learningto, younger2001}) demonstrated how LSTMs (Long Short Term Memory networks) can be used for meta-learning. We find LSTMs useful in our meta learning approach and their use is well documented in similar approaches, such as  (\citet{andrychowicz2016learning,Ravi2017OptimizationAA}), who present a method for optimally applying an incoming gradient signal to parameters by using an LSTM. Further novel work on non-traditional learning rules can by found in the field of feedback alignment (\citet{lillicrap2014random, nkl2016direct}), where the authors show that backpropagation can still work with an altered weight matrix in the backward pass, instead of using the transpose of the weight matrix as would be derived in traditional backpropagation. Our work is to the largest degree inspired by the original MAML paper (\citet{finn2017modelagnostic}, and follow-up works \citet{li2017metasgd, antoniou2018train}), and our training regime can be seen as an enhancement of it. An approach implementing hebbian learning was recently meta-learned through end-to-end backpropagation (\citet{miconi2020backpropamine}).  A black-box evolutionary approach was recently explored in AutoML Zero (\citet{real2020automl}). We find such a clean-slate approach desirable and we believe it could be used to meta-learn MPLP-based models too. Our method was inspired by the recent Neural CA work (\citet{mordvintsev2020growing}) - end-to-end differentiable cellular automata can be seen as a special case of a locally connected computation graph being trained with meta-learning. The recent talk by \citet{aguera2019} demonstrating bacteria-inspired lifeforms and evolutionary approaches to meta-learning was also a direct inspiration to our work.

Methods that fulfill meta-learning objectives such as ours can often be used for solving few-shot learning tasks. \citet{triantafillou2019metadataset} provides a valuable overview of few-shot learning and introduces a large dataset for evaluating architectures. Recently, it is becoming apparent that feature extraction as a pre-processing step may be very beneficial to scaling and improving performance on few-shot learning tasks (\citet{Gidaris_2018, qi2017lowshot, chen2019closer}). We suspect the our approach can be directly applied on learned representations of data. See Section~\ref{sec:discussion} for more details.

Recent work in meta-learning has independently yielded approaches that share some key properties with our work. We hereby highlight the similarities and differences. \citet{metz2018metalearning} trains meta-learning unsupervised learning rules through backpropagation. During the inner update, they backpropagate information through a learned matrix, and instead of receiving a gradient of a loss, each synapse is responsible for accumulating useful statistics in an unsupervised fashion. In summary, the information sent across layers is scalar for each synapse, as opposed to our work, and furthermore, they do not use losses in the inner loop. \citet{gregor2020finding} uses k-dimensional messages as a communication protocol and it is conceptually very similar to our work. However, they specifically set out to solve the task of learning to remember, and do not attempt to generalize to a swap-in replacements to gradient based approaches as we do. \citet{bertens2019network} also uses k-dimensional message passing and a variant of an LSTM/GRU as an underlying building block. Unlike our approach, they rely on evolution based learning and do not set out to find a replacement for gradient-based approaches for traditional NNs.

Continual learning is another field where we believe there are opportunities to make use of our framework.  Historically, meta-learning approaches were not common in continual learning, see \citet{parisi2018continual} for an overview of the field. However, we are aware of some meta-learning work occurring in recent years. Follow the meta leader (FTML) (\citet{finn2019online}) adapts MAML to work in a continual learning scenario; Online
aware Meta-learning (OML) (\citet{javed2019metalearning}) meta-learns sparse representations aiding in continual learning; A Neuromodulated Meta-Learning Algorithm (ANML) (\citet{beaulieu2020learning}) meta-learns architectures resistant to catastrophic forgetting by adding a gating mechanism. In Section~\ref{sec:discussion}, we discuss how MPLP could be applied to Continual learning, and where we see possible limitations/drawbacks.

\textbf{Graph Neural Networks.} From a Graph Neural Networks (GNN) perspective, our framework could be considered an example of Message Passing Neural Network (\citet{gilmer2017neural}). However, GNNs are generally not applied to meta-learning, instead are used to ingest graph-structured data. We recommend \citet{Wu_2020} for a survey on GNNs.

%% file: model.tex
\section{Model}\label{sec:model}

\begin{figure}
\centering
\includegraphics[width=14cm]{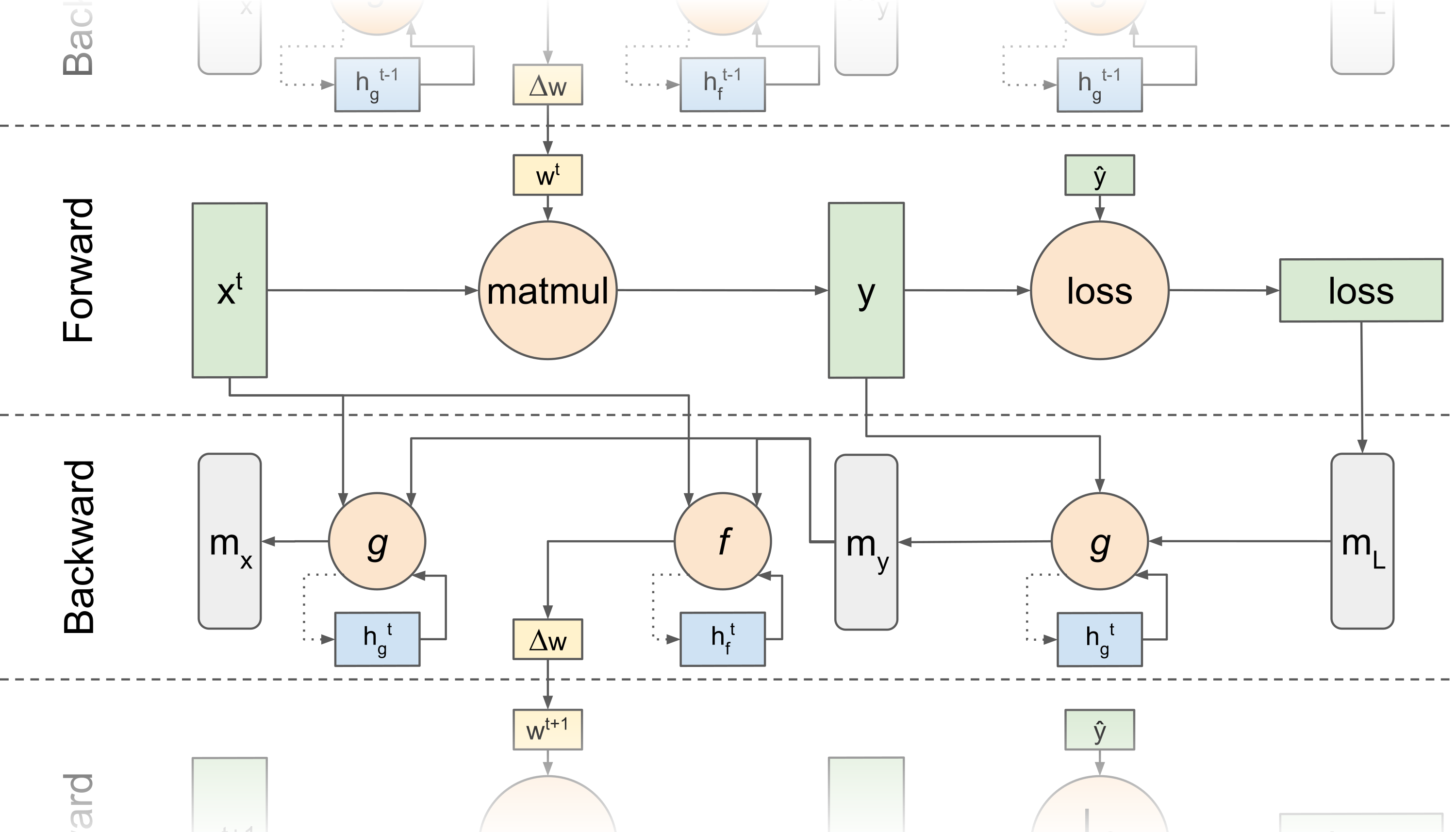}
   \caption{Diagram of simple nodes being trained with MPLP, representing a 1-dimensional matrix multiplication and a loss. $f$ and $g$ are the weight update and message generating networks, respectively. They each maintain their own recurrent hidden state.}
\label{fig:mplp_schema}
\end{figure}

We designed our approach around a directed computational multidigraph. We refer to computations happening within and across nodes as \textit{arrows}. Arrows are responsible for computing messages and updating internal states of nodes. For a more detailed explanation of the actual implementation, we point the reader to Appendix~\ref{app:comp_graph}.

 In this work, we focus on specialising our models to mimic traditional feed-forward neural networks forward and backward passes in supervised learning scenarios.

\subsection{Feed-forward Neural Networks}
A typical feed-forward neural network $f$ is composed of two data pathways: a forward pass, where given some input $x$ and parameters $\theta$ we compute $y = f(x, \theta)$, and a backward pass, where a given loss $\mathcal{L}$ and some stored intermediate data, typically computed after a forward pass, are used to compute a gradient-based update on $\theta$. Our computational model can be used in a forward/backward routine for architectures equivalent to traditional NN, where the forward pass is effectively unchanged. However, instead of relying on gradients for the backward pass, we meta-learn a MPLP.

We now discuss the logical implementation of both forward and backward passes by focusing on the local interactions that each node has. For instance, instead of describing procedures in terms of a weight matrix $W$, we will show what happens to each individual synapse $w_{ij}$ locally. The actual implementation computes updates in bulks; Figure~\ref{fig:mplp_schema} shows a higher level example of dataflow.

In the \textbf{forward pass}, we construct nodes for each operation happening in affine transforms, activations, and loss functions. Each weight and bias is stored in the state of the respective node. We define a forward arrow to compute one-dimensional outputs from one-dimensional inputs, just like what occurs in a traditional forward pass of neural networks; for instance, a forward pass of a weight is $y_{ij} = w_{ij} \cdot x_i$, and stores $x_i$ for future use in the backward pass. $y_{ij}$ is further aggregated to $y_j = \sum_k y_{kj}$. The result of a matrix multiplication can therefore be constructed by these local operations to obtain the more familiar $y = Wx$. Likewise, a sigmoid layer $y = \sigma(x)$ can be deconstructed locally for each scalar $x_i \in x$ and $y_i \in y$ as follows: $y_i = \sigma(x_i)$, storing $x_i$ for further use in the backward pass.

In the \textbf{backward pass}, every node computes a message to send back, given its stored forward input, the message being passed from the successive layer, and any internal states. We refer to this function as the \textit{message passing} rule, or $g$. For a given node indexed $i$, we compute the message $m_i$ from message $m_j$, forward input $x_i$ and internal state $h_i$, which consists of any parameters specific to the node's forward pass as well as any recurrent hidden state.
\begin{equation}
    m_{i} = g(m_j, x_i, h_i)
\end{equation}
Operations such as the loss function and activation functions have no parameters, so the internal states of these nodes only store the pre-activation inputs\footnotemark and a hidden recurrent state.

\footnotetext{Some nodes take more inputs. This is purely done to reduce the total number of nodes necessary for complex operations. For instance, Softmax nodes store every intermediate operation result and the common denominator across all inputs.}

Seeding the backward pass requires passing an initial message to the loss node. We typically pass multidimensional messages between nodes, however we treat the message being externally passed into the loss node as a special case of one-dimensional message:

\begin{equation}
    m_{loss_i} = [loss_i]
\end{equation}

The output message of a loss node being passed backwards is multidimensional.

Nodes representing parameterized operations, such as an affine transform, further undergo a \textit{weight update rule} defined by $f$ during the backwards pass.

\begin{equation}
    \Delta w_{ij} = f(m_j, x_i, h_i)
\end{equation}
For an input with a batch of size B, at step $t$, the update is the average of the update computed over every element in the batch.
\begin{equation}
    w^{t+1}_{ij} = w^t_{ij} + \frac{\sum\limits^B \Delta w^t_{ij}}{|B|}
\end{equation}

Given a traditional dense layer with a weight matrix of cardinality $N \cdot M$, and forward and backward arrows for each individual weight and bias, it is evident some of the messages have to be replicated and others aggregated. Each of the $M$ neurons will have their own bias, which will send the same message to each of its $N$ inputs, as they are connected to $N$ incoming nodes and associated weights. This is not unlike how the backwards flowing gradient is sent along all paths. Similarly, the output of the backward pass of a dense layer is expected to be $N$ messages, one for each input. We accomplish that by averaging the messages being passed backwards to each input.
\begin{equation}
    m_i = \frac{\sum\limits^K m_{ik}}{K}
\end{equation}
Other strategies may be worth exploring, such as summing backward flowing messages, or even more complex and stateful aggregation such as feeding messages into an RNN.

\textbf{Stateful and stateless learners.} 
We parameterize the backwards arrows, $f$ and $g$, using deep neural networks. We experiment with both stateful and stateless implementations of these arrows. We consider the backward network to be stateless in the case when there is not recurrent hidden state between messages or updates being computed - the backwards pass has no memory of previous iterations. To implement the backwards pass with a memory, we incorporate a hidden state at each node and implement $f$ and $g$ as using a Gated Recurrent Unit (GRU). We observed that a deep network is beneficial for the computation of the next state update:

\begin{equation}
    u = \sigma (x * W_{ux} + c^t * W_{uc} + b_u)
\end{equation}
\begin{equation}
    r = \sigma (x * W_{rx} + c^t * W_{rc} + b_r)
\end{equation}
\begin{equation}
    c^{t+1} = u \cdot c^t + (1 - u) \cdot \tanh{(MLP_x(x)+ MLP_c(c^t \cdot r) + b_n)}
\end{equation}

The MLP have two hidden layers of size 80 and 40 respectively with ReLu activations. We consider some of the carry states as output messages, and weight update for weights and biases.
In Experiment~\ref{sec:sinusoid_prior}, we used a stateless version of the above, where we use a MLP for $f$, and a MLP for $g$, both of them with two hidden layers of size 80 and 40 respectively with ReLu activations, and a tanh activation for the final layer. We did not explore the possible space of architectures further.

\textbf{Normalization.} Meta-learning is notoriously hard to train. We found initializations to be critical for successful training. In particular, for any $f$ and $g$ inputs, it is evident that input messages, carry states, inputs to the forward pass, and optional weights have all different means and magnitudes. We mitigated this problem by standardizing these inputs individually over initial minibatches. The mean and standard deviation from these initial minibatches are recorded, and reused thorough the network's lifetime to standardize subsequent batches. Outputs need to be translated and scaled as well: the outputs of $f$ are inherently bounded in (-1, 1) by the tanh activation function, while the typical magnitude and standard deviation of weights and biases is usually orders of magnitude smaller. This empirically results in too large weight and bias updates during early training, rendering meta-training particularly unstable. Moreover, having a mean output significantly different from zero causes suboptimal training. To mitigate these problems, before starting training, for each dense layer, we store the output mean of $f_W$ and $f_b$ (respectively for the weights and biases) and use them to have initial mean zero outputs. Furthermore, we scale outputs down to be at most $0.2$ times the standard deviation of the weight matrix\footnotemark. We keep all these normalization variables fixed during meta-training, except for the output scaling, which we meta-train as we observed it to be beneficial for fast adaptation. We believe other standardization techniques such as batch normalization may prove fruitful, but we do not explore them in this work.

\footnotetext{This means that weights and biases of the same dense layer are initialized to have $f$s output similar magnitudes. This is of particular importance, since we zero-initialize biases, and therefore we could not compute their initial standard deviations.}

\textbf{Parameter sharing.} Across all our experiments, all weight nodes for a given weight matrix, and all bias nodes for a given bias vector, share $f$ and $g$\footnotemark. Likewise, all nodes for a given activation function, and all nodes for a given loss, share $g$. In addition, we experimented with two more configurations: sharing f and g across all layers of the same kind (so we have one f and one g for all weight nodes in the network, and another set of f and g for all bias nodes), and sharing f, g and the standardization parameters described above.

\footnotetext{Weights and biases \textbf{do not} share $f$ and $g$ among themselves. The same reasoning applies with loss functions and activations: a cross-entropy loss node \textbf{does not} share $g$ with a ReLU node.}

\subsubsection{Training regime}

\textbf{Cross-validation loss.} We compute a cross-validation loss after performing k-steps of adaptation, similarly to the meta-training procedure used in MAML.

Let us call $F$ and $G$ the forward pass and backward pass functions of a neural network. In our use case, $G$ can be interpreted as MPLP. The neural network is parameterized by $\theta$ and MPLP is parameterized by $\phi$. Let $\mathcal{L}$ be an arbitrary loss function. We adapt the neural network on a training set $(x^{(t)}, \hat{y}^{(t)})$ and keep a heldout set $(x^{(e)}, \hat{y}^{(e)})$.

We update $\theta$ with the training set:
\begin{equation}\label{eq:l1}
    y^{(ti)} = F(x^{(ti)}, \theta_i)
\end{equation}
\begin{equation}\label{eq:l2}
    \theta_{i+1} = G(\mathcal{L}(y^{(ti)}, \hat{y}^{(ti)}), \theta_i, \phi)
\end{equation}
This is repeated k times. Afterwards, we compute a cross-validation loss with the heldout set:
\begin{equation}
    \mathcal{L}_{cv} = \mathcal{L}(F(x^{(e)}, \theta_k), \hat{y}^{(e)})
\end{equation}
While we do not pass hidden states to the equations above, you can assume every function ingests and returns hidden states.

\textbf{Hint losses.} In addition to the cross-validation loss described above, we have observed it is beneficial to add a \textit{hint loss}:
at each step of the inner loop, \textit{after} Equations \ref{eq:l1} and \ref{eq:l2}, we compute a loss on the same training data just observed:
\begin{equation}
    \mathcal{L}_{hi} = \mathcal{L}(F(x^{(ti)}, \theta_{i+1}), \hat{y}^{(ti)})
\end{equation}
This meta-loss is scaled such that the sum of all hint losses across all the inner loop steps has the same magnitude as the final cross-validation loss. We find better convergence by weighting the total hint loss to be larger than the global meta-loss. We suspect such an intermediate loss produces a smoother local loss landscape, but its inherent greediness may also hamper our model in finding an optimal convergence path. In Experiment~\ref{sec:sinusoid_prior} we show what can happen when we do not add a hint loss.

\textbf{Outer batches.} We meta-learn by accumulating gradients across multiple tasks, generally referred to as outer batches. In this work, tasks refer to: (a) a selection of amplitude and phase for sinusoidals, (b) a randomly initialized network to optimize for either sinusoidals and MNIST. Regarding the latter, for all experiments except Experiment~\ref{sec:sinusoid_prior}, we train MPLP able to adapt any randomly initialized networks. We do so by initializing\footnotemark a pool of networks before training $[\theta^1_0, \theta^2_0, ... \theta^n_0]$, and using a different randomly sampled prior $\theta^i_0$ for every outer batch. Even very small pools are generally sufficient for MPLP to generalize on unseen random initializations. However, a pool of size 1 would have stateful learners overfit to it.

\footnotetext{We initialize weights sampled from a normal distribution of mean 0 and standard deviation of 0.05. We zero-initialize biases.}

We meta-learn $\phi$ through Adam and normalize the gradients for each optimizer parameter. We further describe training setups specific to each experiment in their relative sections.

%% file: sinusoid.tex
\subsection{Sinusoidal fitting}

Following the example in MAML, we show convergence on tasks drawn from a family of sinusoidal functions. We sample a set of sinusoidal tasks for each training step, where a sinusoidal task is defined as a regression on samples from the underlying function $ y(x) = A \sin (x + p)$ , with $A \in [0.1, 0.5]$ and $p \in [0, \pi]$. We sample inner batches with $x \in [-5, 5]$. For an inner training loop consisting of k steps, we sample k inner batches. We show convergence of our approach both when trained from a randomly initialized prior, and when combined with MAML to learn prior weights of the optimizee network. We use 8-dimensional messages.

\subsubsection{Randomly initialized priors}\label{sec:sinusoid_random}

In this experiment, we train MPLP to generalize to adapt for arbitrary sinusoidals and arbitrary networks initializations. Thus, for each outer batch, we sample a random sinusoidal task and a randomly initialized network prior. The optimizee network is a traditional deep neural network with two hidden layers of 20 units each, and ReLu activations on the hidden layers. We use stateful learners for MPLP and do not share any parameters across layers.

\textbf{Training.} We perform 5-step learning with an inner batch size of 10. We use an outer batch size of 4. We use a cross-validation loss at the end of the 5-step learning, and a hint loss at every step. We use an L2 loss for both cross-validation and hint losses. The L2 loss is also the final node of the network (that is, the loss in the initial message in the backward stage is L2).

\input{drawings/sin_10_5_comparison.tex}

Figure~\ref{fig:sin_10_5_comparison} shows a comparison between MPLP and Adam\footnotemark. The MPLP optimizer was trained for \textasciitilde 40,000 steps. The optimized MPLP is able to fit the task, and it vastly outperforms what Adam can achieve in 5 steps.

\footnotetext{We use out-of-the-box Adam parameters, except for the learning rate, that we fine-tuned to be 0.01, since it gave the best results.}

Training MPLP showed a great variance in its convergence results. Anecdotally, we observed a trend of obtaining better and more reliable results the larger number of parameters MPLP has.  However, all parameter sharing configurations can, albeit seldom, converge to successful MPLP similar to what shown in Figure~\ref{fig:sin_10_5_comparison}. We give more details in Appendix~\ref{app:sinusoid_ablation}.

\subsubsection{Learning with a prior}\label{sec:sinusoid_prior}

In this experiment, we show how it is possible to meta learn the priors (as in MAML) and MPLP jointly. To showcase different properties of MPLP from what observed in Experiment~\ref{sec:sinusoid_random}, we choose to use a stateless learning algorithm, perform 2-shot learning, and use only a cross-validation loss at the end (i.e. no hint loss). We also do not share any parameters across layers.

\input{drawings/sin_maml_comparison.tex}

Figure \ref{fig:sin_maml_comparison} showcases how adding MPLP to MAML can drastically change the behaviour of the learner. While both models successfully solve the task, MAML runs are all always approximating a sinusoidal with priors. Moreover, since they are forced to follow the gradient at every step, it is no surprise they increasingly approximate a sinusoidal. Using MPLP, instead, the learning algorithm is not restricted to performing incremental improvements at every step. In this case, this results in nothing like sinusoidal approximation until the last step. We also verified that the learned MPLP is tightly coupled with the learned priors, as it would not succeed with a randomly initialized network.
Meta-training on this experiment will yield different MPLPs and henceforth different visual results. Therefore, what we show in Figure~\ref{fig:sin_maml_comparison} is just one example of many possibilities.

%% file: drawings/sin_10_5_comparison.tex
\begin{figure}
\centering
\begin{tikzpicture}[
 image/.style = {text width=0.45\textwidth, 
                 inner sep=0pt, outer sep=0pt},
node distance = 0mm and 0mm
                        ] 
\node [image, label={MPLP}] (frame1)
    {\includegraphics[width=\linewidth]{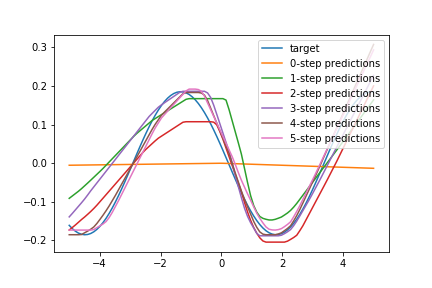}};
\node [image,right=of frame1, label={Adam}] (frame2) 
    {\includegraphics[width=\linewidth]{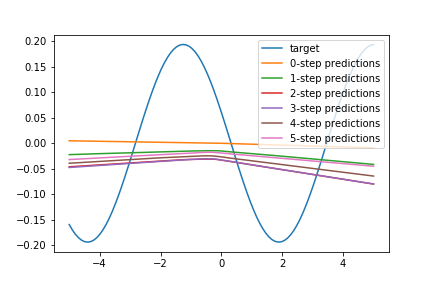}};
\node[image,below=of frame1] (frame3)
    {\includegraphics[width=\linewidth]{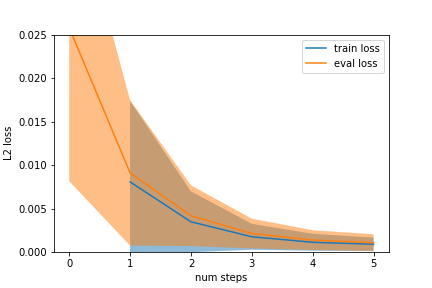}};
\node[image,right=of frame3] (frame4)
    {\includegraphics[width=\linewidth]{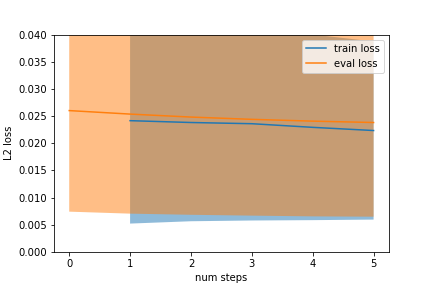}};
\end{tikzpicture}
   \caption{Comparison of MPLP (left column) learning with Adam (learning rate of 0.01) (right column). The top row shows a typical example run for the evaluation of 5-step training with inner batch size of 10. The bottom row shows means and standard deviations of the losses over 5 steps, across 100 runs; the eval loss is the L2 loss averaged across the entire domain, the train loss is the L2 loss average across all training points observed so far.}
\label{fig:sin_10_5_comparison}
\end{figure}

%% file: drawings/sin_maml_comparison.tex
\begin{figure}
\centering
\begin{tikzpicture}[
 image/.style = {text width=0.45\textwidth, 
                 inner sep=0pt, outer sep=0pt},
node distance = 1mm and 1mm
                        ] 
\node [image, label={MPLP \& MAML}] (frame1)
    {\includegraphics[width=\linewidth]{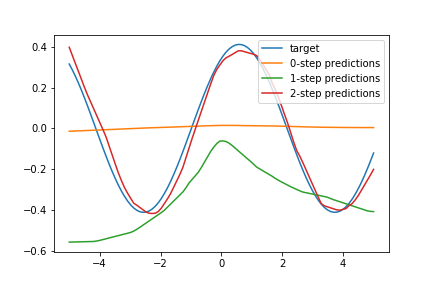}};
\node [image,right=of frame1, label={MAML}] (frame2) 
    {\includegraphics[width=\linewidth]{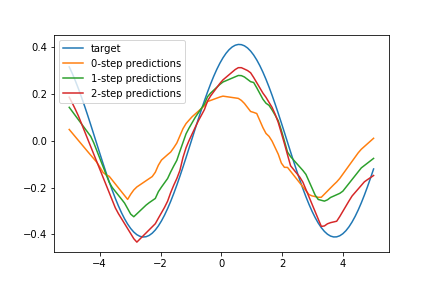}};
\end{tikzpicture}
   \caption{Comparison of models trained with MPLP and MAML (left), and standalone MAML (right).}
\label{fig:sin_maml_comparison}
\end{figure}

%% file: mnist.tex
\subsection{MNIST classification}

The goal of this experiment is to learn a MPLP capable of generalizing to arbitrary network initializations to adapt on MNIST. To do so, we perform a 20-step training with inner batch size of 8 on a scaled-down (12x12) MNIST. We scale-down MNIST mostly for performance reasons, but this allows us to evaluate our trained learners on a full-scale MNIST afterwards. We also standardize the inputs by computing mean and standard deviation on the train dataset. The architecture used is a (144,50,10) network with sigmoid activation for the hidden layer, a softmax for the final layer, and a cross-entropy loss node afterwards. Every task in the outer batch is initialized with a different network prior. We used a stateful learner with a message size of 4 as a compromise between quality and computational efficiency.

\textbf{Parameter sharing.} We share the learners for all the dense layers (all weights share one pair of $f$ and $g$, all biases share one pair of $f$ and $g$), and keep normalization parameters layer-specific. We could not manage to achieve comparable results when sharing normalization parameters across layers, and we defer more robust explorations to future work.

\textbf{Training regime.} We use an outer batch size of 1 and meta-learn for 5000 steps. We do so for computational constraints and it is by no means optimal. We compute a hint loss for every step and a cross-validation loss after 20 steps. The losses used are cross-entropy.

\begin{figure}
\centering
\includegraphics[width=10cm]{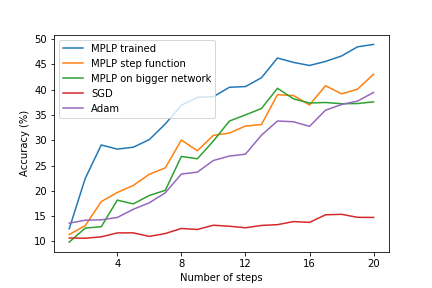}
   \caption{Evalautions of MPLP, SGD and Adam on the MNIST test set. For all models except "MPLP on bigger network", we run this on a 12x12 sized MNIST dataset. Every plot point is computed by averaging 100 test points for every of the 100 runs. Table~\ref{tab:mnist_acc} shows standard deviations on the last step.}
\label{fig:mnist_comparisons}
\end{figure}

\begin{table}
  \caption{MNIST Accuracy at 20 steps}
  \label{tab:mnist_acc}
  \centering
  \begin{tabular}{lll}
    \toprule
    \cmidrule(r){1-2}
    Model     & Architecture & Accuracy @ 20 (\%)     \\
    \midrule
    MPLP & (144,50,10), Sigmoid & $ 48.95 \pm 6.35 $     \\
    MPLP (transfer) & (144,50,10), Step & $ 43.07 \pm 7.36 $     \\
    MPLP (transfer) & (784,100,10), Sigmoid & $ 37.60 \pm 9.60 $     \\
    SGD & (144,50,10), Sigmoid & $ 14.74 \pm 5.80 $     \\
    Adam & (144,50,10), Sigmoid & $ 39.46 \pm 9.83 $     \\
    \bottomrule
  \end{tabular}
\end{table}

\textbf{Results.} Figure~\ref{fig:mnist_comparisons} and Table~\ref{tab:mnist_acc} show evaluations on the test set, with randomly initialized priors, of our MPLP compared to Adam and SGD\footnotemark, averaged over 100 runs.
\footnotetext{Here as well, we fine-tune the learning rate of both Adam and SGD to 0.01 and 0.1 respectively.}
Most evaluations are performed on the scaled-down 12x12 MNIST on a (144,50,10) architecture. The exception is for the "MPLP on bigger network", where we transfer the learnt parameters to a larger network of size (784,100,10). We can see how MPLP outperforms Adam and SGD, on average, for every step. We also experimented with replacing the sigmoid function in the hidden layer with a step function, since the two are closely related. While a step function would only backpropagate zero gradients with traditional means, we are able to train a network with it seamlessly. Finally, we ported the trained MPLP to a larger network of dimensions (784,100,10), with a sigmoid as hidden layer, and evaluated its performance on a 28x28 MNIST. Even if we did not share any standardizers, we observe how the model is still able to train meaningfully.

%% file: outro.tex
\section{Discussion}\label{sec:discussion}

In this article, we introduced MPLP and applied it to fairly small feedforward neural networks with the aim of inquiring whether this is a viable generalization to gradient-based approaches. In this section, we want to share our thoughts and discuss some of the promising results, as well as some eye-catching hardships we encountered.

Overall, for few-shot learning, a properly normalized MPLP appears to be a promising path to pursue. All the experiments explored suggest that meta-learning MPLP for few-shot tasks should work for small-sized networks. If we wanted to scale this up to larger sized networks, this algorithm can quickly become overly expensive. For instance, some of our early experiments implemented convolutions for stateless learners, and we saw how that can very quickly become too computationally expensive. We therefore hypothesize a scaling up might require some significant architectural enhancement, or, more likely, finding a compromise between locality and scalability. One other viable path could be rethinking traditional NN blocks, making them inherently more powerful through MPLP, decreasing architectural depth.

In this work, we never passed gradients in the backward step. This may not necessarily be optimal for real-world experiments, and we observed some good arguments in favour of passing gradients whenever possible: A typical meta-training regime tends to start with the learners not knowing what to do at all. This results in some early loss plateau, that the learners have to get out of. This plateau gets harder to escape with badly initialized networks. We hypothesize this to be less of a problem when also passing the gradient of the loss, since the model could very easily escape such plateaus by just scaling the received gradients.

We also showed a toy example on how a MPLP trained for sigmoids transfers to step functions too. While we would not recommend reading too much into that specific result, we could imagine some meta-meta-training regimes that could improve MPLP to eventually get completely rid of differentiability requirements, while transfering well to non differentiable operations.

MPLP can be applied to arbitrary graphs. As long as the building blocks are end-to-end differentiable, MPLP can be learned efficiently. We expect that different tasks may benefit from nontraditional NN blocks, and recommend trying MPLP. The code is open sourced on \href{https://github.com/google-research/self-organising-systems/tree/master/mplp}{https://github.com/google-research/self-organising-systems/tree/master/mplp}. 

Continual learning is also a great fit for meta-learning and MPLP: we can train learners that follow cross-validation losses, or losses that discourage forgetfulness. However, we observed one drawback with our loss-driven approach: meta-learning MPLP suffers from all the well known problems of training NNs. In particular, we should consider MPLP as composed by small neural networks, and they can quickly fall to bad local minima. We observed these problems in nearly every experiment we performed, including some we did not include in this work. One solution may be to overparameterize MPLP, but this is likely inefficient. One other solution is to use the right set of losses to hint MPLP to go towards a proper direction. This is why we introduced hint losses. Hint losses are even more useful for longer time series such as 50 steps or more. Catastrophic forgetting avoidance would as well require a hint loss for remembering over time. However, finding the right hint loss for the right task is not straightforward. We consider this the biggest limitation we observed in our experiments.

In summary, MPLP can be a viable way to overcome some of the problems gradient-based approaches have: adding a cross-validation loss can drive the learners to optimize for generalizing results, and not overfit; catastrophic forgetting avoidance can as well be encoded in the meta-loss used; the learned MPLP does not require differentiable architectures, if we are able to transfer it effectively; vanishing and exploding gradients may be less of an issue, given that the messages passed do not necessarily need to decrease/increase in magnitude to express certain wishes of changes to be performed downstream.

More broadly, one can learn a MPLP on any end-to-end differentiable configuration of agents. This may lead to exploring novel architectures, where each agent is inherently more complex and powerful than traditional operations, or where computations happen asynchronously and distributedly. MPLP could also be used for devising implicit reward signals in continual learning or reinforcement learning scenarios.

%% file: appendix.tex
\appendix
\appendixpage

\section{Computational Graph details}\label{app:comp_graph}

For ease of implementation we designed our approach around a directed computational multidigraph. Nodes in our computational graph are stateful and store arbitrary data structures which we refer to as "state". Arrows (directed edges) in our graph represent a series of operations to be executed, each taking as input the state of the source node, potentially modifying this state or part of it, and then constructing and passing a message into the state of the destination node. This passed message contains any information the source node wants to pass on to the destination node. The received message can be instantly processed by the destination node by executing a successive arrow computation, or stored for future usage. It is important to note that the mapping of a node's received messages to the inputs of its outgoing arrows is defined during the construction of the graph. Likewise, the order of execution of the edges is arbitrary but in the specific case of traditional networks we emulate the forward-backward paradigm - executing forward arrows in sequence until the final message being passed is the loss value, then executing backward arrows in the reverse sequence to update the parameters stored in the state. By property of a multidigraph, two nodes can have more than one edge defined between them, as is the case with the aforementioned forward and backward arrows. A node can additionally have reflexive arrows, allowing updates to its own states without receiving any external inputs. An example of how we utilize internal states in our experiments is storing weights or bias values, Gated Recurrent Unit (GRU)/LSTM hidden states, or a node's embedding.

Such a framework allows for a variety of possible architectures for arbitrary computation and communication within a network.

\section{Sinusoidal Ablation study}\label{app:sinusoid_ablation}

We performed ablations on message size and amount of shared parameters (see Section~\ref{sec:model} for the meaning of the possible configurations). Unfortunately, we observed a great amount of variance for each experiment, and it was too computationally prohibitive to perform several repetitions for each instance. What therefore follows is anecdotal evidence.
Ablating the message size (we tried 1, 4, and 8), we observed that training converges faster and the final result is better with a larger message size. Ablating the amount of shared parameters, we observed the models to converge faster with no shared parameters, but the end results are all comparable even if we share all parameters. A note on reliable results: the least powerful the models, the more likely they are to get stuck in some local minima, and we observed every model to get stuck, in some runs; the most common problem we observed was it getting stuck to classifying a center arc only, while regressing to classifying a constant value outside of the center.